\RequirePackage{etoolbox}
\csdef{input@path}{%
 {sty/}
 {img/}
}%
\csgdef{bibdir}{bib/}

\documentclass[ba]{imsart}
\pubyear{0000}
\volume{00}
\issue{0}
\doi{0000}
\firstpage{1}
\lastpage{1}

\usepackage{amsfonts}
\usepackage{blkarray}
\usepackage{amsthm}
\usepackage{amsmath}
\usepackage{natbib}
\usepackage{hyperref}
\usepackage{graphicx}

\usepackage{algorithm2e}

\newcommand{\mb}[1]{\mathbf{#1}}

\startlocaldefs
\numberwithin{equation}{section}
\theoremstyle{plain}

\endlocaldefs

\begin{document}

\begin{frontmatter}
\title{Bayesian Inference over the Stiefel Manifold via the Givens Representation}
\runtitle{Bayesian Inference via the Givens Representation}

\begin{aug}
\author{\fnms{Arya A. Pourzanjani}\thanksref{addr1}\ead[label=e1]{arya@ucsb.edu}},
\author{\fnms{Richard M. Jiang}\thanksref{addr1}},
\author{\fnms{Brian Mitchell}\thanksref{addr1}},
\author{\fnms{Paul J. Atzberger}\thanksref{addr2}}
\and
\author{\fnms{Linda R. Petzold} \thanksref{addr1}
}

\runauthor{Pourzanjani et al.}

\address[addr1]{Computer Science Department, University of California, Santa Barbara
    \printead{e1} 
}

\address[addr2]{Mathematics Department, University of California, Santa Barbara
}

\end{aug}

\begin{abstract}
We introduce an approach based on the Givens representation for posterior inference in statistical models with orthogonal matrix parameters, such as factor models and probabilistic principal component analysis (PPCA).  We show how the Givens representation can be used to develop practical methods for transforming densities over the Stiefel manifold into densities over subsets of Euclidean space.  We show how to deal with issues arising from the topology of the Stiefel manifold and how to inexpensively compute the change-of-measure terms.  We introduce an auxiliary parameter approach that limits the impact of topological issues.  We provide both analysis of our methods and numerical examples demonstrating the effectiveness of the approach.  We also discuss how our Givens representation can be used to define general classes of distributions over the space of orthogonal matrices.  We then give demonstrations on several examples showing how the Givens approach performs in practice in comparison with other methods.  
\end{abstract}



\end{frontmatter}

\section{Introduction}
Statistical models parameterized in terms of orthogonal matrices are ubiquitous, particularly in the treatment of multivariate data. This class of models includes certain multivariate time series models \citep{brockwell2002introduction}, factor models \citep{johnson2004multivariate}, and many developed probabilistic dimensionality reduction models such as Probabilistic PCA (PPCA),  Exponential Family PPCA (BXPCA), mixture of PPCA \citep{ghahramani1996algorithm}, and Canonical Correlation Analysis (CCA) \citep[Chapt.~12.5]{murphy2012machine}. These models were traditionally used in fields such as psychology \citep{ford1986application}, but more recently are gaining traction in other diverse applications including biology \citep{hamelryck2006sampling}, finance \citep{lee2007bayesian}, materials science \citep{oh20172d}, and robotics \citep{lu1997robot}.

\noindent Despite their ubiquity, routine and flexible posterior inference in Bayesian models with orthogonal matrix parameters remains a challenge. Given a specified posterior density function, modern probabilistic programming frameworks such as Stan, Edward, and PyMC3~\citep{carpenter2016stan,tran2016edward,salvatier2016probabilistic} can automatically generate samples from the associated posterior distribution with no further input from the user. Unfortunately, these current frameworks do not offer support for density functions specified in terms of orthogonal matrices. While innovative methods have been introduced to handle such densities, these approaches are often specialized to specific probability distributions or require re-implementations of the underlying inference methods~\citep{hoff2009simulation,brubaker2012family,byrne2013geodesic,holbrook2016bayesian}. 

\noindent An appealing alternative approach to this problem is to parameterize the space of orthogonal matrices, i.e. the Stiefel manifold, in terms of unconstrained Euclidean parameters, then use this parameterization to transform a posterior density on the space of orthogonal matrices to a posterior density on Euclidean space. The resulting density could then be used to sample from the posterior distribution using a probabilistic programming framework such as Stan. 

\noindent While appealing, the transformation approach can pose its own challenges related to the change in measure, topology, and parameterization. While there are many possible parameterizations of orthogonal matrices \citep{anderson1987generation, shepard2015representation}, we seek smooth continuously differentiable representations that can be used readily in inference methods such as Hamiltonian Monte Carlo (HMC). It is also important to consider for transformed random variables the change-of-measure adjustment term which needs to be computable efficiently. For the Stiefel manifold, the topology also can pose issues when mapping to Euclidean space. Finally, in practice, we also seek representations that have an intuitive interpretation helpful in statistical modeling and specifying prior distributions.

\noindent We introduce an approach to posterior inference for statistical models with orthogonal matrix parameters based on Givens representation of orthogonal matrices~\citep{shepard2015representation}. Our approach addresses several of the concerns of using a transformation-based approach. We derive an analytic change-of-measure adjustment term that is efficient to compute in practice. We also address two topological issues that occur when using the Givens representation to transform densities. The first issue has to do with multi-modalities in the transformed density that are introduced by the transform. We resolve it by a parameter expansion scheme. The second issue has to do with how the Givens representation pathologically transforms densities near ``poles" of the Stiefel manifold. We present both theory and numerical results showing how this plays a minimal role in practice. We also discuss how the Givens representation can be used to define new distributions over orthogonal matrices that are useful for statistical modeling. Our approach enables the application of sampling methods such as the No-U-Turn Sampler (NUTS)~\citep{hoffman2014no} to arbitrary densities specified in terms of orthogonal matrices. Our approach is relatively easy to implement and has the advantage that current sampling algorithms and software frameworks can be leveraged without needing significant modifications or re-implementations. We expect this work to help facilitate practitioners' ability to build and prototype rapidly complex probabilistic models with orthogonal matrix parameters in probabilistic frameworks, such as Stan, Edward, or PyMC3. 

\noindent We organize the paper as follows. In Section~\ref{related}, we discuss existing methods for Bayesian inference over the Stiefel manifold and the difficulty in implementing these methods in existing Bayesian software frameworks. In Section~\ref{Givens} we describe the Givens representation by first introducing the Givens reduction algorithm and then connecting it to a geometric perspective of the Stiefel manifold. We try to provide an approachable intuition to the transform. We then describe how to practically apply the Givens representation to sample from posterior densities that are specified in terms of orthogonal matrices in Section \ref{implementation}. In Section~\ref{examples} we present results for statistical examples demonstrating in practice our Givens representation approach and how it compares with other methods.  

\section{Related Work} \label{related}
While several innovative methods have been proposed for sampling distributions over the Stiefel manifold, many are either specialized in the types of distributions they can sample or require significant modifications of the underlying inference methods. This poses challenges for their adoption and use in practice, especially within current probabilistic programming frameworks used by practitioners.

\noindent In \cite{hoff2009simulation}, a Gibbs sampling approach is introduced for densities specified in terms of orthogonal matrices when the orthogonal matrix parameter, conditioned on all other model parameters, follows a distribution in the Bingham-von Mises-Fisher family. In practice, this limits the flexibility of this approach to a specific class of models and may not offer the same sampling efficiency as modern algorithms such as HMC. 

\noindent Recently, HMC-based methods have been proposed for sampling distributions over the Stiefel manifold and handling constraints~\citep{brubaker2012family,byrne2013geodesic,holbrook2016bayesian}. The work of \cite{brubaker2012family} proposes a modified HMC algorithm which uses an update rule for constrained parameters based on the symplectic SHAKE integrator \citep{leimkuhler2004simulating}. Specifically, for unconstrained parameters, the method uses a standard Leapfrog update rule. For constrained parameters, the method first takes a Leapfrog step which usually moves the parameter to a value that does not obey constraints. The method then uses Newton's method to ``project" the parameter value back down to the manifold where the desired constraints are satisfied. In \citet{byrne2013geodesic} and \citet{holbrook2016bayesian} a separate HMC update rule is introduced to deal with constrained parameters. These works utilize analytic results, and the matrix exponential function to update the parameters in such a way that guarantees constraints are still satisfied in the embedded matrix coordinates. More precisely, they use the fact that analytic solutions for the geodesic equations on the Stiefel manifold in the embedded coordinates are known. This gives rise to their Geodesic Monte Carlo (GMC) algorithm. While this does provide a mathematically elegant approach, in practice, the matrix exponential and use of separate update rules in GMC makes the algorithm difficult to implement in general.

\noindent Unfortunately, these recently introduced HMC-based methods methods rely on specialized HMC update rules which can make them difficult to implement within probabilistic programming frameworks and to tune. In particular, they sample distributions with orthogonal matrix parameters by using different HMC update rules for constrained and unconstrained parameters, requiring additional book-keeping in software to know which update rules to use on which parameter. Unfortunately, many probabilistic programming languages do not keep track of this as they treat parameters agnostically by transforming to an unconstrained space. Without the automatic tuning of inference algorithm parameters offered in common probabilistic programming frameworks, these algorithms must be manually tuned in practice as they usually do not offer an algorithm to choose tuning parameters automatically. Furthermore, the specialization of these methods to HMC makes them difficult to generalize to other inference algorithms based on variational inference (VI) or optimization, which would be more straight-forward with a transformation based approach.

\noindent We also mention the interesting recent work of \cite{jauch2018random}, in which they also try to address related difficulties using a transformation-based approach, but based instead on the Cayley Transform.  As in our Givens representation approach, their approach also faces the challenges of topological difficulties of mapping the Stiefel manifold to Euclidean space and in computing efficiently the change-of-measure term. We present here for the Givens representation ways to grapple with these issues which we expect are applicable generally to such methods.

\section{The Givens Representation of Orthogonal Matrices} \label{Givens}
We discuss the the Givens reduction algorithm of numerical analysis and describe the connection between the algorithm and the geometric aspects of the Stiefel manifold. We then describe the Givens representation of orthogonal matrices.

\subsection{Givens Rotations and Reductions} \label{givens_reduction}
Given any $n \times p$ matrix, $A$, the Givens reduction algorithm is a numerical algorithm for finding the $QR$-factorization of $A$, i.e. an $n\times p$ orthogonal matrix $Q$ and an upper-triangular, $p \times p$, matrix $R$ such that $A = QR$. The algorithm works by successively applying a series of Givens rotation matrices so as to ``zero-out" the elements $\{A_{ij} : i > j \}$ of $A$. These Givens rotation matrices are simply $n \times n$ matrices, $R_{ij}(\theta_{ij})$, that take the form of an identity matrix except for the $(i,i)$ and $(j,j)$ positions which are replaced by $\cos \theta_{ij}$, and the $(i,j)$ and $(j,i)$ positions which are replaced by $-\sin \theta_{ij}$ and $\sin \theta_{ij}$ respectively.

\begin{figure}[h]
\centering
\vspace{.1in}
\includegraphics[width=0.4\textwidth]{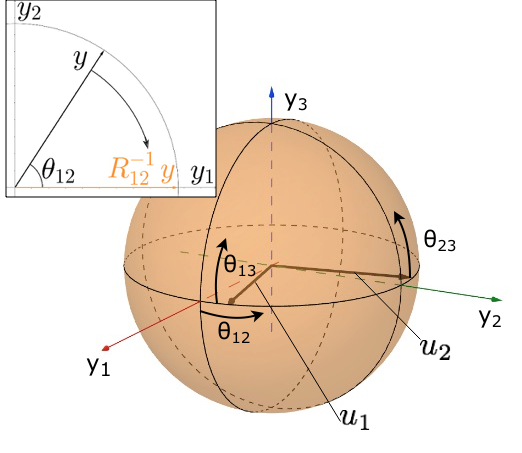}
\vspace{.05in}
\caption{(Inset) Givens rotations can be used to rotate a vector so as to eliminate its component in a certain direction. (Main Figure) A $p$-frame on the Stiefel manifold can be visualized as a set of rigidly connected orthogonal basis vectors, $u_1$ and $u_2$, shown here in black. One can move about the Stiefel manifold and describe any $p$-frame by simultaneously applying rotation matrices of a prescribed angle to these basis vectors. Applying the rotation matrix $R_{12}(\theta_{12})$ corresponds to rotating the two basis vectors together in the (1,2)-plane, which by our convention is the $(x,y)$-plane. Similarly, simultaneously applying $R_{13}(\theta_{13})$ corresponds to a rotation of the 2-frame in the $(1,3)$ or $(x,z)$-plane, while $R_{23}(\theta_{23})$ corresponds to rotating $u_2$ about $u_1$.}
\label{fig:StiefelGeom}
\end{figure}

\noindent When applied to a vector, $R_{ij}(\theta_{ij})$ has the effect of rotating the vector counter-clockwise in the $(i,j)$-plane, while leaving other elements fixed. Intuitively, its inverse, $R_{ij}^{-1}(\theta_{ij})$, has the same effect, but clockwise. Thus one can ``zero-out" the $j$th element, $u_j$, of a vector $u$, by first using the $\arctan$ function to find the angle, $\theta_{ij}$ formed in the $(i,j)$-plane by $u_i$ and $u_j$, and then multiplying by the matrix $R_{ij}^{-1}(\theta_{ij})$ (Figure \ref{fig:StiefelGeom}, inset).

\begin{figure}[h]
\centering
\vspace{.1in}
\includegraphics[width=1.0\textwidth]{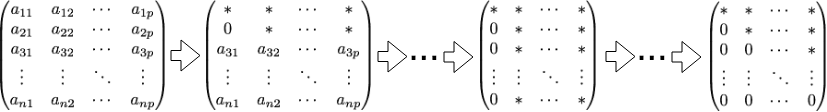}
\vspace{.05in}
\caption{The Givens reduction eliminates lower diagonal elements of an $n \times p$ matrix one column at a time. Because each rotation, $R_{ij}(\theta_{ij})$, only affects rows $i$ and $j$, previously zeroed out elements do not change.}
\label{fig:givens_reduction}
\end{figure}

\noindent In the Givens reduction algorithm, these rotation matrices are applied one-by-one to $A$ in this way to zero-out all elements below the $(i,i)$ elements of the matrix. First, all elements in the first column below the first row are eliminated by successively applying the rotation matrices $R_{12}^{-1}(\theta_{12}), R_{13}^{-1}(\theta_{13}), \cdots, R_{1n}^{-1}(\theta_{1n})$  (Figure \ref{fig:givens_reduction}). Because multiplication by $R_{ij}(\theta_{ij})$ only affects elements $i$ and $j$ of a vector, once the $j$th element is zeroed out, the subsequent rotations, $R_{13}^{-1}(\theta_{13}), \cdots, R_{1n}^{-1}(\theta_{1n})$, will leave the initial changes unaffected. Similarly, once the first column of $A$ is zeroed out below the first element, the subsequent rotations, which do not involve the first element will leave the first column unaffected. The rotations  $R_{23}^{-1}(\theta_{23}), \cdots, R_{2n}^{-1}(\theta_{2n})$ can thus be applied to zero out the second column, while leaving the first column unaffected. This results in the upper triangular matrix

\begin{equation}
\label{eq:givens_reduction}
R_* := \underbrace{R_{pn}^{-1}(\theta_{pn}) \cdots R_{p,p+1}^{-1}(\theta_{p,p+1})  \cdots R_{2n}^{-1}(\theta_{2n}) \cdots R_{23}^{-1}(\theta_{23}) \cdots R_{1n}^{-1}(\theta_{1n}) \cdots R_{12}^{-1}(\theta_{12})}_{Q_*^{-1}} A.
\end{equation}

\noindent Crucially, the product of rotations, which we call $Q_*^{-1}$, is orthogonal since it is simply the product of rotation matrices which are themselves orthogonal. Thus its inverse can be applied to both sides of Equation \ref{eq:givens_reduction} to obtain

\begin{equation}
Q_* R_* = A.
\end{equation}

\noindent The familiar $QR$ form can be obtained by setting $Q$ equal to the first $p$ columns of $Q_*$ and setting $R$ equal to the first $p$ rows of $R_*$. The Givens reduction is summarized in Algorithm \ref{alg:givens_reduction}.

\begin{algorithm}[h]
\SetAlgoLined
\KwIn{$A$}
\KwResult{$Q,R$} 
 $Q_*^{-1} = I$
  $R_* = A$
 
 \For{$i$ in 1:p}{
 \For{$j$ in (i+1):n}{
    
    $\theta_{ij} = \arctan(Y[j,i]/Y[i,i])$
    
    $Q_*^{-1} = R_{ij}^{-1}(\theta_{ij}) Q_*^{-1}$
    
    $R_* = R_{ij}^{-1}(\theta_{ij}) R_*$

 }
 }
return $Q_*[,1:p], R_*[1:p,1:p]$
\\
\caption{Psuedo-code for the Givens reduction algorithm for obtaining the $QR$ factorization of a matrix $A$.}
 \label{alg:givens_reduction}
\end{algorithm}

\subsection{The Geometry of Orthogonal Matrices}\label{givens_stiefel_geometry}
The elements of the Stiefel manifold, $V_{p,n}$,  are known as $p$-frames. A $p$-frame is an orthogonal set of $p$ $n$-dimensional unit-length vectors, where $p \le n$. $p$-frames naturally correspond to $n \times p$ orthogonal matrices which can be used to define the Stiefel manifold succinctly as

\begin{equation}
V_{p,n} := \{Y \in \mathbb{R}^{n \times p}: Y^TY = I \}.
\end{equation}

\noindent Geometrically, an element of the Stiefel manifold can be pictured as a set of orthogonal, unit-length vectors that are rigidly connected to one another. A simple case is $V_{1,3}$, which consists of a single vector, $u_1$, on the unit sphere. This vector can be represented by two polar coordinates that we naturally think of as longitude and latitude, but can also be thought of simply as subsequent rotations of the standard basis vector $e_1 := (1,0,0)^T$ in the $(x,y)$ and $(x,z)$ planes, which we refer to as the $(1,2)$ and $(1,3)$ planes for generality. In mathematical terms, $u_1$ can be represented as $u_1 = R_{12}(\theta_{12}) R_{13}(\theta_{13}) e_1$~(Figure \ref{fig:StiefelGeom}). 

\noindent Continuing with our geometric interpretation, $V_{2,3}$ can be pictured as a vector in $V_{1,3}$ that has a second orthogonal vector, $u_2$, that is rigidly attached to it as it moves about the unit sphere. Because this second vector is constrained to be orthogonal to the first, its position can be described by a single rotation about the first vector. Thus elements of $V_{2,3}$ can be represented by three angles: two angles, $\theta_{12}$ and $\theta_{13}$, that represent how much to rotate the first vector, and a third angle, $\theta_{23}$ that controls how much the second vector is rotated about the first (Figure \ref{fig:StiefelGeom}). Mathematically this can be represented as the $3 \times 2$ orthogonal matrix $R_{12}(\theta_{12}) R_{13}(\theta_{13}) R_{23}(\theta_{23}) (e_1, e_2)$.

\noindent Although elements of the Stiefel manifold can be represented by $n \times p$ matrices, their inherent dimension is less than $np$ because of the constraints that the matrices must satisfy. The first column must satisfy a single constraint: the unit-length constraint. The second column must satisfy two constraints: not only must it be unit length, but it must also be orthogonal to the first column. The third column must additionally be orthogonal to the second column, giving it a total of three constraints. Continuing in this way reveals the inherent dimensionality of the Stiefel manifold to be

\begin{equation}
\label{eq:stiefel_dimension}
d := np - 1 - 2- \cdots p  = np - \frac{p(p+1)}{2}.
\end{equation}

\subsection{Obtaining the Givens Representation}\label{givens_representation_introduction}
The Givens reduction applied to an orthogonal matrix gives rise to a representation of the Stiefel manifold that generalizes the intuitive geometric interpretation described above. When applied to an $n \times p$ orthogonal matrix $Y$, the Givens reduction yields 

\begin{equation}
\label{eq:inverse_givens_representation}
R_{pn}^{-1}(\theta_{pn}) \cdots R_{p,p+1}^{-1}(\theta_{p,p+1})  \cdots R_{2n}^{-1}(\theta_{2n}) \cdots R_{23}^{-1}(\theta_{23}) \cdots R_{1n}^{-1}(\theta_{1n}) \cdots R_{12}^{-1}(\theta_{12}) Y = I_{n,p}
\end{equation}

\noindent where $I_{n,p}$ is defined to be the first $p$ columns of the $n \times n$ identity matrix, i.e. the matrix consisting of the first $p$ standard basis vectors $e_1, \cdots, e_p$. The first $n-1$ rotations transform the first column into $e_1$, since it zeros out all elements below the first and the orthogonal rotations do not affect the length of the vector which by hypothesis is unit length. Similarly, the next $n-2$ rotations will leave the length of the second column and its orthogonality to the first column intact because again, the rotation matrices are orthogonal. Therefore, because the second column must be zero below its second element, it must be $e_2$ after these $n-2$ rotations are applied. Continuing in this way explains the relationship in Equation \ref{eq:inverse_givens_representation}.

\noindent Because $Y$ was taken to be an arbitrary orthogonal matrix, it is clear from Equation \ref{eq:inverse_givens_representation} that any orthogonal matrix $Y$ can be factored as

\begin{equation}
\label{eq:givens_representation}
Y = R_{12}(\theta_{12}) \cdots R_{1n}(\theta_{1n})  \cdots R_{23}(\theta_{23}) \cdots R_{2n}(\theta_{2n}) \cdots R_{p,p+1}(\theta_{p,p+1}) \cdots R_{pn}(\theta_{pn}) I_{n,p}.
\end{equation}

\noindent Defining $\Theta := (\theta_{12} \cdots \theta_{1n} \cdots \theta_{23} \cdots \theta_{2n} \theta_{p,p+1} \cdots \theta_{pn})$ we can consider any orthogonal matrix as a function, $Y(\Theta)$, of these angles, effectively parameterizing the Stiefel manifold and yielding the Givens representation. The Givens representation is a smooth representation with respect to the angles $\Theta$ \citep{shepard2015representation}, and lines up with our geometric insight discussed in the previous subsection. We also note that the number of angles in the Givens representation corresponds exactly to the inherent dimensionality, $d$, of the Stiefel manifold.

\section{Using the Givens Representation to Sample Distributions Over the Stiefel Manifold} \label{implementation}
Using the Givens representation in practice to sample distributions over the Stiefel manifold requires solving several practical challenges. In addition to the standard change-of-measure term required in any transformation of a random variable, care must be taken to address certain pathologies that occur due to the differing topologies of the Stiefel manifold and Euclidean space. We further describe these challenges and how we overcome them in practice. We also briefly remark on how the Givens representation can be leveraged to define new and useful distributions over the Stiefel manifold. We conclude the section by describing how the computation of the Givens representation scales in theory, particularly in comparison to GMC.

\subsection{Transformation of Measure Under the Givens Representation}\label{measureGivens}
\begin{figure}[h]
\centering
\vspace{.1in}
\includegraphics[width=0.5\textwidth]{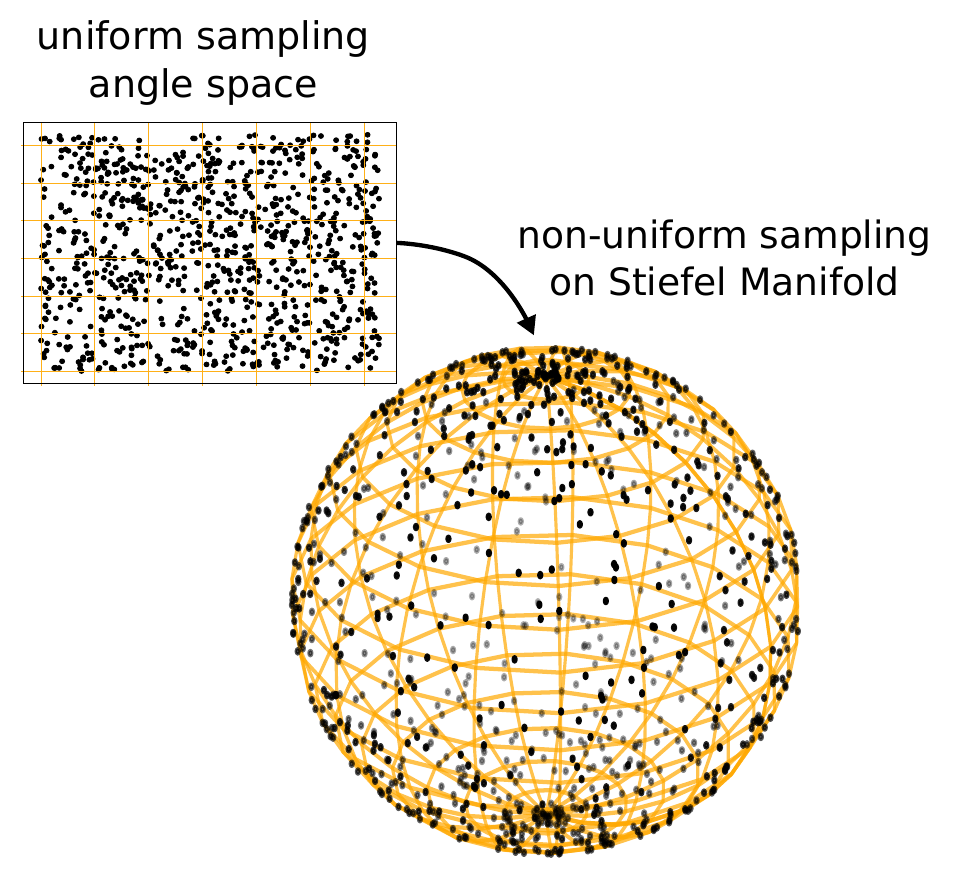}
\vspace{.05in}
\caption{Uniform sampling in the Givens representation coordinates does not necessarily lead to uniform sampling over the Stiefel manifold without the proper measure adjustment term. Under the mapping, regions near the pole are shrunk to regions on the sphere with little area, as opposed to regions near to the equator which the transform maps to much larger areas on the sphere. Intuitively, the change-of-measure term quantifies this proportion of shrinkage in area.}
\label{fig:AreaForm}
\end{figure}

As is usual in any transformation of random variables, care must be taken to include an extra term in the transformed density to account for a change-of-measure under the transformation. Formally, for a posterior density over orthogonal matrices that takes the form $p_Y(y)$, the proper density over the transformed random variable, $\Theta(Y)$, takes the form $p_\Theta(\theta) = p_{Y}(Y(\theta)) |J_{Y(\Theta)}(\theta)|$ \citep{keener2011theoretical}. Intuitively, this extra term accounts for how probability measures are distorted by the transformation (Figure \ref{fig:AreaForm}). Usually this term is calculated by simply taking the determinant of the Jacobian of the transformation. Unfortunately, the Givens representation, $Y(\Theta)$, is a map from a space of dimension $d := np - p(p+1)/2$ to a space of dimension $np$. Thus its Jacobian is non-square and the determinant of the Jacobian is undefined.

\noindent One way to compute the change-of-measure term analogous to the Jacobian determinant is to appeal to the algebra of differential forms. Denote the product of $n \times n$ rotation matrices in the Givens representation by $G$, i.e. 

\begin{equation}
G := R_{12}(\theta_{12}) \cdots R_{1n}(\theta_{1n})  \cdots R_{23}(\theta_{23}) \cdots R_{pn}(\theta_{pn}) \cdots R_{p,p+1}(\theta_{p,p+1}) \cdots R_{pn}(\theta_{pn}),
\end{equation}

\noindent and denote its $j$th column by $G_j$. \cite{muirhead2009aspects} shows that the proper measure for a signed surface element of $V_{p,n}$ is given by the absolute value of the differential form

\begin{equation}
\label{eq:WedgeForm}
\bigwedge_{i=1}^p \bigwedge_{j=i+1}^n G_j^T\, dY_i.
\end{equation}

\noindent Letting $J_{Y_i(\Theta)}(\Theta)$ be the Jacobian of the $i$th column of $Y$ with respect to the angle coordinates of the Givens representation, this differential form can be written in the coordinates of the Givens representation as

\begin{equation}
\label{eq:WedgeForm_givens}
\bigwedge_{i=1}^p \bigwedge_{j=i+1}^n G_j^T\, J_{Y_i(\Theta)}(\Theta) d\Theta.
\end{equation}

\noindent Because this is a wedge product of $d$ $d$-dimensional elements, Equation \ref{eq:WedgeForm_givens} can be conveniently written as the determinant of the $d \times d$ matrix

\begin{equation}
\label{eq:measure_matrix_form}
\begin{pmatrix}
G_{2:n}^T J_{Y_1(\Theta)}(\Theta)\\
G_{3:n}^T J_{Y_2(\Theta)}(\Theta)\\
\vdots\\
G_{p:n}^T J_{Y_p(\Theta)}(\Theta),
\end{pmatrix}
\end{equation}

\noindent where the $G_{k:l}$ denote columns $k$ through $l$ of $G$. As we show in the Appendix, this term, which would otherwise be expensive to compute, can be analytically simplified to the following simple-to-compute product whose absolute value serves as our measure adjustment term:

\begin{equation}
\label{eq:final_change_of_measure}
J_{Y(\Theta)}(\Theta) = \prod_{i=1}^p \prod_{j=i+1}^n \cos^{j-i-1} \theta_{ij}.
\end{equation}

\subsection{Implementation of Angle Coordinates}
Two issues related to the topology of the Stiefel manifold arise when using the Givens representation to map densities over the Stiefel manifold to densities over Euclidean space. Let $\theta_{12}, \theta_{23}, \cdots \theta_{p,p+1}$ range from $-\pi$ to $\pi$. We refer to these specific coordinates as the latitudinal coordinates to evoke the analogy for the simple spherical case. Similarly, let the remaining coordinates range from $-\pi/2$ to $\pi/2$. We refer to these coordinates as longitudinal coordinates. Formally, this choice of intervals defines a coordinate chart from Euclidean space to the Stiefel manifold, i.e. the correspondence mapping between these two spaces.

\begin{figure}[h]
\centering
\vspace{.1in}
\includegraphics[width=0.5\textwidth]{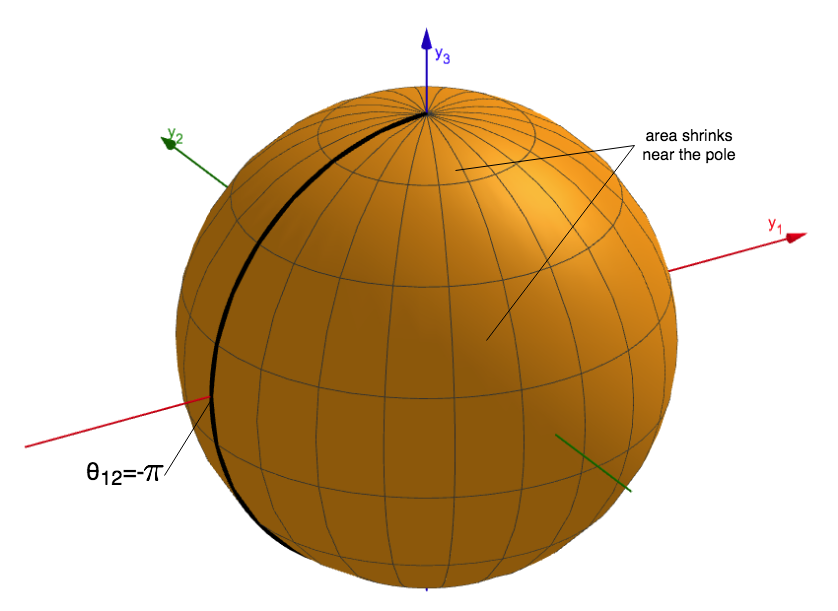}
\vspace{.05in}
\caption{The angular coordinate chart has an infinitesimal sliver of measure zero lying at $\theta_{12} = -\pi = \pi$ that separates two otherwise connected parts of the sphere. Trajectories $Y(t)$ over the Stiefel manifold that cross this sliver have no equivalent representation, $\Theta(t)$, in the coordinates of the Givens representation. This can become particularly problematic when there is significant probability mass on both sides of the sliver. The grid over the sphere reveals how the Givens representation maps areas that are the same size in the $\Theta$ coordinates to smaller and smaller regions on the sphere the closer they are to the poles. Thus the measure adjustment term (Equation \ref{eq:final_change_of_measure}), which measures how the transform changes the area of these infinitesimal regions, goes to zero near the poles, making the transformed density at these points zero regardless of whether they were assigned to zero by the original density.}
\label{fig:pathologies}
\end{figure}

\noindent While this choice of intervals allows us to represent almost the entire Stiefel manifold in the Givens representation, because the topology of these two space differ, certain connectedness properties of the Stiefel manifold can not be accurately represented in the Givens representation. For example, when representing $V_{1,3}$ in Euclidean space using the Givens representation, contiguous regions of the manifold on either side of the sliver corresponding to $\theta_{12} = \pi$ are disconnected (Figure \ref{fig:pathologies}). As we show, this could lead to samples that are not representative of the correct distribution when applying sampling methods such as HMC. To address this, we introduce auxiliary parameters to the Givens representation to better represent the connectedness of the Stiefel manifold in Euclidean space.

\noindent In addition to these disconnected regions, the coordinate chart will also contain singularities where the measure adjustment term (Equation \ref{eq:final_change_of_measure}) approaches zero near the ``poles'' of the Stiefel manifold i.e. where the longitudinal coordinates equal $-\pi/2$ or $\pi/2$. This means that a finite density over the Stiefel manifold that is transformed to a density over Euclidean space using the Givens representation will equal zero at the poles of the Stiefel manifold regardless of whether the original density assigns zero to those points. On the sphere, this happens at the North and South poles where the longitudinal coordinates become exactly $-\pi/2$ or $\pi/2$ (Figure \ref{fig:pathologies}). In practice, this prevents algorithms such as HMC from obtaining samples in a small region near these points even when there is positive probability mass in these regions under the original density. The reason is that the acceptance ratio used by algorithms such as HMC will always equal zero at these points for finite densities. Thus proposals at these points will always be rejected. Because of finite numerical precision, this also holds true for points on the Stiefel manifold that are numerically near these poles. While in theory this would necessitate a change in coordinate charts near these regions, fortunately, in practice these pathological regions generally have a negligible effect. We present both theoretical analysis and empirical numerical experiments showing the minimal impacts of these regions in numerical estimates of expectations obtained using the Givens representation.

\subsubsection{Auxiliary Parameters for Addressing Connectedness}
\begin{figure}[h]
\centering
\vspace{.1in}
\includegraphics[width=0.6\textwidth]{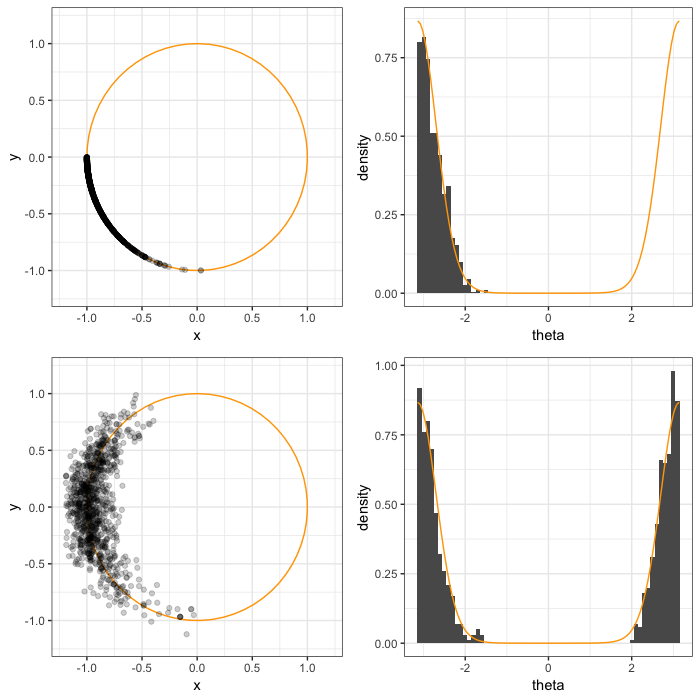}
\vspace{.05in}
\caption{(Upper) 1,000 samples from a Von Mises distribution with parameters $\mu = -\pi$ and $\kappa = 5$ sampled over the space $\theta \in [-\pi, \pi]$ using Stan. Most of the mass of the distribution is concentrated at the ends of the interval while little mass is concentrated towards the middle. Because these two ends of the interval are disconnected in this representation, the sampler gets ``stuck" in the mode corresponding to the $-\pi$ side of the interval rather than the $\pi$ side. (Lower) 1,000 samples from the equivalent distribution sampled over the $(x,y)$-space. By introducing an auxiliary coordinate, one can effectively replicate the topology of a circle, effectively ``wrapping" the two ends of the interval so that the sampler avoids getting stuck in one region.}
\label{fig:donut}
\end{figure}

\noindent  Simply allowing the latitudinal coordinates of the Givens representation to range from $-\pi$ to $\pi$ leaves regions of parameter space that should otherwise be connected, disconnected in the Givens representation. For a distribution over the Stiefel manifold that is not sufficiently concentrated away from these disconnected regions, this can lead to highly non-representative samples when naively applying the Givens representation for sampling (Figure \ref{fig:donut}, upper).

\noindent To handle this multimodality introduced by reparameterizing in terms of Givens angles, we introduce for each angle parameter, $\theta_{ij}$, an independent auxiliary parameter, $r_{ij}$. We then transform the density to sample over the $x_{ij},y_{ij}$-space via the transform $x_{ij} = r_{ij} \cos \theta_{ij}$ and $y_{ij} = r_{ij} \sin \theta_{ij}$. In the transformed space, the two ends of the interval are connected, producing samples that are distributed more evenly across the two disconnected regions (Figure \ref{fig:donut}, lower). Formally, we assign to $r$ a marginal distribution with density $p_r(r)$ so that $\theta$ and $r$ are independent and the marginal distribution of $\theta$ is left untouched by the introduction of the auxiliary parameter. This leads to the joint density $p_{\theta, r}(\theta, r) = p_\theta(\theta) p_r(r)$ which we then transform to the equivalent density over the unconstrained $(x,y)$-space by the simple change-of-variables formula between two-dimensional polar coordinates and two-dimensional Euclidean coordinates:

\begin{eqnarray}
p_{x, y}(x,y) &=& p_{\theta, r}(\mathrm{arctan} \left(y/x), \sqrt{x^2 + y^2} \right) \frac{1}{r}.
\end{eqnarray}

\noindent This again leaves the marginal distribution of $\theta$ unaffected, however, in the new space, paths $\theta(t)$ that cross the region of parameter space at $\theta = \pi$ can actually be represented. In practice, we set  $p_r(r)$ to a normal density with mean one and standard deviation 0.1. Although $r_{ij}$ does not necessarily need to be set to this particular distribution to achieve the correct marginal distribution over $\theta$, this choice helps to avoid the region of parameter space where $r = 0$ and the transformed density is ill-defined.

\subsubsection{Transformation of Densities Near the Poles}
\noindent  Even with the usual change of variables formula and the measure adjustment term (Equation \ref{eq:final_change_of_measure}), a finite density over the Stiefel manifold that is transformed to a density over Euclidean space using the Givens representation will not be completely equivalent to the original density. In particular, when any of the longitudinal angles has absolute value equal to $\pi/2$, Equation \ref{eq:final_change_of_measure} will equal zero. Thus the transformed density will be zero at these points even when the original density is non-zero there. Because of finite numerical precision, in practice this creates a region of the Stiefel manifold that cannot be sampled by algorithms such as HMC, despite having a positive probability mass under the original distribution. Specifically, because a computed numerical density will be zero at values numerically near the poles, the acceptance ratio in HMC will always be zero at these points so that proposals in the region will always be rejected. This effectively blocks off a portion of parameter space by limiting all longitudinal angles to the region $[-\pi/2 + \epsilon, \pi/2 - \epsilon]$ where $\epsilon$ is a small value on the order of numerical precision. While a change in coordinate chart could be utilized, we show that in practice the exceedingly small volume of this region mitigates the effect of these regions on numerical samplers.

\noindent For general $n$ and $p$, the volume of this blocked off region is $\mathcal{O}(p \epsilon^2)$. First note that the uniform density over $V_{p,n}$ in the Givens representation is simply a constant times the absolute value of Equation \ref{eq:final_change_of_measure}. However, since this density factors into a product of independent terms, the longitudinal angles are independent of one another. The probability that at least one longitudinal angle falls within the $\epsilon$-region is thus equal to the sum of the individual probabilities of each angle falling within the region plus higher order terms. Each of these individual probabilities is proportional to $\cos^{j-i-1} \theta_{ij}$, which for small $\epsilon$ can be bounded by $\epsilon^{j-i-1}$ over the interval $[\pi/2 - \epsilon, \pi/2]$. Thus the probability of falling within the $\epsilon$-region is bounded by a constant times the following quantity:

\begin{equation}
\label{eq:bound_on_epsilon_vol}
\sum_{i=1}^p \sum_{j=i+2}^n 2 \int_{\pi/2-\epsilon}^{\pi/2} \epsilon^{j-i-1} d\theta_{ij} = \sum_{i=1}^p \sum_{j=i+2}^n 2 \epsilon^{j-i} = \sum_{i=1}^p \mathcal{O}(\epsilon^2) = \mathcal{O}(p \epsilon^2).
\end{equation}

\noindent Because this quantity falls off with the square of $\epsilon$, even for modestly small $\epsilon$, the probability of a uniformly sampled point falling within the $\epsilon$-region is small. Empirical results further illustrate this. For various values of $n, p,$ and $\epsilon$ we drew 100,000 samples uniformly form the Stiefel manifold by sampling the elements of an $n \times p$ matrix from a standard normal distribution, then taking the QR factorization of this matrix, a common technique for uniformly sampling the Stiefel manifold \citep{muirhead2009aspects}. We then took these samples, converted them into their Givens representation, and calculated the number of samples that had any longitudinal angle within the region $[-\pi/2, -\pi/2+\epsilon]$ or the region $[\pi/2-\epsilon, \pi/2]$. The results are closely explained by Equation \ref{eq:bound_on_epsilon_vol}. In particular, the proportion of samples that fell within this region does not change much for fixed $p$ and increasing $n$. Furthermore, the proportion increases linearly with $p$, and it decreases quadratically with $\epsilon$ (Table \ref{tab:uniform_epsilon_region}).  

\begin{table*}
\begin{tabular}{|cc||ccccc|}
\hline
$p$ & $n$  & $\epsilon = 0.1$ & $\epsilon = 0.05$ & $\epsilon = 0.025$ & $\epsilon = 0.0125$ & $\epsilon = 1e-5$\\
\hline
\hline
1 & 10 & 490 & 114 & 22 & 4 & 0\\
1 & 20 & 499 & 118 & 25 & 4 & 0\\
1 & 50 & 570 & 148 & 32 & 6 & 0 \\
\hline
3 & 10 & 1,612 & 381 & 79  & 15 & 0\\
3 & 20 & 1,665 & 398 & 78 & 19 & 0\\
3 & 50 & 1,712 & 416 & 100  & 24 & 0\\
\hline
10 & 10 & 4,260 & 1,071 & 258 & 59 & 0\\
10 & 20 & 5,342 & 1,336 & 357 & 91 & 0\\
10 & 50 & 5,266 & 1,368 & 334 & 90 & 0 \\
\hline
\end{tabular}
\caption{The number of uniform samples out of 100,000 that fell within the $\epsilon$ region for various values of $n, p,$ and $\epsilon$. Samples are taken uniformly from the Stiefel manifold using the QR factorization method. As the theoretical bound suggests, the number of samples falling in this region increases modestly for fixed $p$ and increasing $n$. It increases linearly with $p$, and it decreases quadratically with $\epsilon$. In particular, whenever $\epsilon$ is halved, the number of samples falling within the region decreases by about a fourth. We also note that for $\epsilon = 1e-5$, the value we used for most of our experiments, the number of samples falling within the $\epsilon$ region was zero for all settings.}
\label{tab:uniform_epsilon_region}
\end{table*}

\noindent For non-uniform distributions with a probability density $p(\Theta)$ that is finite in the $\epsilon$-region, the probability of any of the longitudinal angles falling within the $\epsilon$ region can again be bounded by a constant times $\mathcal{O}(p \epsilon^2)$. We took 100,000 samples from the von Mises Fisher distribution over $V_{1,3}$ with parameters $\mu = (0,0,1)$ and $\kappa = 1, 10, 100$, and 1000 using the simulation method of \citet{wood1994simulation} as implemented in the R package Rfast. For fixed $\kappa$ the probability of a sample falling in the $\epsilon$-region drops off with the square of $\epsilon$ as the bound would suggest. This holds true even when probability mass is highly concentrated near these regions (Table \ref{tab:vmf_epsilon_region}), although for highly distributions that are highly concentrated near the poles we advise users to conduct a similar sensitivity analysis for their particular case.  

\begin{table*}
\begin{tabular}{|c||ccccc|}
\hline
$\kappa$  & $\epsilon = 0.1$ & $\epsilon = 0.05$ & $\epsilon = 0.025$ & $\epsilon = 0.0125$ & $\epsilon = 1e-5$\\
\hline
\hline
1 & 630 & 163 & 38 & 9 & 0\\
10 & 4,839 & 1,220 & 317 & 81 & 0\\
100 & 39,287 & 11,658 & 3,086 & 764 & 0 \\
1,000  & 99,295 & 71,066 & 26,643 & 7,473 & 0 \\
\hline
\end{tabular}
\caption{The number of samples from a von Mises Fisher distribution with $\mu = (0,0,1)$ and $\kappa = 1, 10, 100$ and 1000 that fell within the $\epsilon$ region for various values of $n, p,$ and $\epsilon$. For each value of $\kappa$, 100,000 total samples were taken. As the theoretical bound suggests, the number of samples that fall within the $\epsilon$ region decreases with the square of $\epsilon$ so that even for a modestly small value of $\epsilon = 1e-5$, none of the 100,000 samples fall within this region even in the highly concentrated case ($\kappa = 1,000$).}
\label{tab:vmf_epsilon_region}
\end{table*}

\begin{figure}[h]
\centering
\vspace{.1in}
\includegraphics[width=0.65\textwidth]{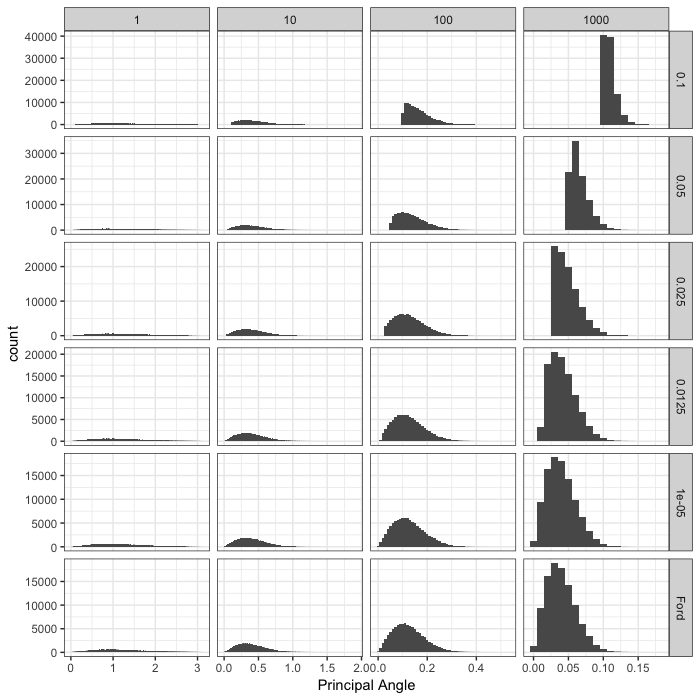}
\vspace{.05in}
\caption{Histograms of the principal angle, $\arccos (\mu^T Y)$, sampled under the von Mises Fisher distribution with $\mu = (0,0,1)$ and $\kappa = 1, 10, 100$ and 1000 using the Givens representation in Stan with various sizes of the $\epsilon$ area and using the method of \citet{wood1994simulation}. For small $\epsilon$ and large $\kappa$, the lack of samples from the $\epsilon$-region in the latter method is evident in the histograms. In particular, despite the large amount of mass near zero, the number of samples is never larger than a bound that is dictated by $\epsilon$. As $\epsilon$ decreases, the bound rapidly becomes negligible because of the quadratic relationship between $\epsilon$ and the volume of the $\epsilon$ area. Thus for $\epsilon = 1e-5$, the histograms of the Givens representation method and the \citet{wood1994simulation} method become indistinguishable.}
\label{fig:epsilon_histogram}
\end{figure}

\noindent Because the probability of samples falling within the $\epsilon$ region falls with the square of $\epsilon$, even for modestly small $\epsilon$ the distribution of derived quantities of $Y(\Theta)$ remains largely unaffected when sampling using the Givens representation with small enough $\epsilon$. We sampled the von Mises Fisher distribution with the same values of $\kappa$ using the Givens representation in Stan but with the longitudinal angles deliberately limited to the interval $[-\pi/2 + \epsilon, -\pi/2+\epsilon]$ with $\epsilon = 0.1, 0.05, 0.025, 0.0125,$ and $1e-5$. We then examined histograms and expectations of the principal angle $\arccos (\mu^T Y)$, which represents the angle between the sample and the direction at the pole which lies directly in the middle of the $\epsilon$-region. For large $\epsilon$ and large $\kappa$ the lack of samples from the $\epsilon$-region is evident when compared to samples using the method of \citet{wood1994simulation} since the sample can not get close enough to $\mu$. However, for any fixed $\kappa$ as $\epsilon$ is decreased, the number of samples that fall within the $\epsilon$-region decreases rapidly as the bound would suggest (Figure \ref{fig:epsilon_histogram}). Table \ref{tab:vmf_epsilon_region_expectations} illustrates this effect numerically using the expectation of the principal angle.

\begin{table*}
\begin{tabular}{|c||ccccc|c|}
\hline
 &   Givens &  &  &  & & Wood\\
\hline
$\kappa$ & $\epsilon = 0.1$ & $\epsilon = 0.05$ & $\epsilon = 0.025$ & $\epsilon = 0.0125$ & $\epsilon = 1e-5$ & \\
\hline
\hline
1 & 1.2027 & 1.2042 & 1.2008 & 1.1995 & 1.1986 & 1.2012\\
10 & 0.4181 & 0.4065 & 0.4031 & 0.4012 & 0.4019 & 0.4015\\
100 & 0.1657 & 0.1377 & 0.1290 & 0.1258 & 0.1261 & 0.1255 \\
1,000 & 0.1092 & 0.0657 & 0.0483 & 0.0422 & 0.0396 & 0.0398\\
\hline
\end{tabular}
\caption{The empirical expectation of the principal angle, $\arccos (\mu^T Y)$, sampled under the von Mises Fisher distribution with $\mu = (0,0,1)$ and $\kappa = 1, 10, 100$ and 1000 using the Givens representation in Stan with various sizes of the $\epsilon$ area and using the method of \citet{wood1994simulation}. As $\epsilon$ decreases, the empirical expectation computed using the Givens representation becomes much closer to those taken via the method of \citet{wood1994simulation}. For small $\kappa$ the expectations do not differ much even for large $\epsilon$ because much less mass concentrates near the $\epsilon$ regions.}
\label{tab:vmf_epsilon_region_expectations}
\end{table*}

\subsection{Specifying Distributions Using the Givens Representation}
So far, we have focused on transforming densities defined in terms of the canonical orthogonal matrix coordinates, $Y$, into densities specified in terms of the angles, $\Theta$, of the Givens representation. However, the angles of the Givens representation can also be used to define new distributions over the Stiefel manifold that may be useful in modeling. In fact, using the intuition described in Section \ref{givens_stiefel_geometry}, the sign and magnitude of the angle $\theta_{ij}$ of the Givens representation roughly corresponds to the sign and magnitude of the $i-j$ element of $Y$. Thus one can create, for example, sparse priors over the Stiefel manifold by placing sparse priors over the angles of the Givens representation.

\noindent \cite{cron2016models} utilize sparsity promoting priors over the coordinates of the Givens representation to produce a prior distribution over covariance matrices that favors sparse matrices. Specifically, they describe a model for multivariate Gaussian observations with an unknown covariance matrix. They parameterize the covariance matrix in terms of its eigen-decomposition $Y \Lambda Y^T$, then parameterize the orthogonal matrix $Y$ using the Givens representation. They then place spike-and-slab mixture priors over the angles of the Givens representation, placing significant prior mass on orthogonal matrices whose Givens angles are mostly zero and thus sparse in the canonical coordinates. They describe a custom reversible jump-based method for sampling the resulting posterior distribution.

\noindent Our Givens representation approach provides a routine and flexible way to sample the posterior distribution associated with this model and other more general models using common probabilistic modeling frameworks. In Section \ref{examples}, we illustrate this with a sparse PPCA example motivated by \cite{cron2016models} that places truncated horseshoe priors on the angles of the Givens representation.


\subsection{Computational Scaling of the Givens Representation}\label{scaling}
The primary computational cost in computing the Givens representation is the series of $d$ $n \times n$ matrix multiplications applied to $I_{n,p}$ in Equation \ref{eq:givens_representation}. Fortunately, unlike dense matrix multiplication, applying a Givens rotation to an $n \times p$ matrix only involves two vector additions of size $p$ (Algorithm \ref{alg:givens}). Thus since $d$ scales on the order of $np$, computation of the Givens representation in aggregate scales as $\mathcal{O}(np^2)$. In comparison, GMC involves an orthogonalization of an $n \times p$ matrix which scales as $\mathcal{O}(np^2)$ and a matrix exponential computation that scales as $\mathcal{O}(p^3)$.

\begin{algorithm}[h]
\SetAlgoLined
\KwIn{$\theta$}
\KwResult{$Y$} 
$Y = I_{n,p}$;
 idx = $d$
 
 \For{$i$ in p:1}{
 \For{$j$ in n:(i+1)}{
    
    $Y_i = \cos(\theta_{\mathrm{idx}}) Y[i,] - \sin(\theta_{\mathrm{idx}}) Y[j,]$
    
     $Y_j = \sin(\theta_{\mathrm{idx}}) Y[i,] + \cos(\theta_{\mathrm{idx}}) Y[j,]$
        
    $Y[i,] = Y_i$
    
    $Y[j,] = Y_j$
    
    $idx = idx - 1$
    
    log density += $(j-i-1) \log \cos \theta_{\mathrm{idx}}$
 }
 }
return $Y$
\\
\caption{Psuedo-code for obtaining the orthogonal matrix $Y$ from the Givens Representation as well as appropriately adjusting the log of the posterior density.}
 \label{alg:givens}
\end{algorithm}

\noindent We note, however, that this comparison is somewhat obfuscated by the gradient of the log probability computation that is required by both methods. When using the Givens representation in a framework such as Stan, this gradient is computed internally using reverse-mode automatic differentiation. Meanwhile, GMC requires user-provided, analytically-known gradients of the log density. These analytically-derived gradients are typically faster to compute than applying reverse-mode automatic differentiation, but this of course limits the types of densities that GMC can be applied to.

\noindent In practice, sampling efficiency will depend on several factors, including the size of the orthogonal matrix being sampled, making it difficult to generally recommend one method over the other in terms of efficiency. For uniform sampling of the Stiefel manifold, we find that GMC scales better when $p$ is much smaller than $n$, whereas the Givens representation scales better when $p$ is large and closer to $n$. We present benchmarks comparing the two methods on orthogonal matrices of various sizes in Section \ref{examples}.

\section{Experiments} \label{examples}
We demonstrate the use of the Givens representation for uniformly sampling the Stiefel manifold as well as several statistical examples. All Givens representation experiments were conducted in Stan using the automatic warm-up and tuning options. For all Stan experiments, we ensured that there were no divergences during post-warmup sampling and that all $\hat{R}$ were $1.01$ or below. Presence of divergences suggests that the sampler may not be visiting areas of the posterior distribution that contain positive mass \citep{betancourt2015hamiltonian}, while $\hat{R}$ tests for convergence of the Markov chain to the stationary distribution \citep{gelman1992inference}. All timing experiments were conducted on a 2016 Macbook Pro.

\subsection{Uniform Sampling on the Stiefel Manifold} \label{scaling_examples}
We sample uniformly from the Stiefel manifold of various sizes to assess the practical scalability of the Givens representation. We compare its sampling efficiency and $\hat{R}$ values to GMC using 500 post-warmup samples from each method (Table \ref{tab:rhat_neff}).  We chose the step size tuning parameter of GMC by manually trying several possible values, then selecting the specific value that produced the highest number of effective samples per second over 500 samples.

\begin{table*}
\begin{tabular}{|cc||cc|cc|}
\hline
& & GMC & & Givens &\\
\hline
$p$ & $n$  & $\hat{R}$ & $n_{\mathrm{eff}}$ & $\hat{R}$ & $n_{\mathrm{eff}}$\\
\hline
\hline
1 & 10 & 1.00 & 231 & 1.00 & 496\\
1 & 100 & 1.00 & 317 & 1.00 & 488\\
1 & 1000 & 1.00 & 238 & 1.00 & 487 \\
\hline
10 & 10 & 1.00 & 408 & 1.00  & 390\\
10 & 100 & 1.00 & 473 & 1.00 & 487\\
10 & 1000 & 1.00 & 454 & 1.00  & 488 \\
\hline
100 & 100 & 1.00 & 484 & 1.00 & 479 \\
\hline
\end{tabular}
\caption{$\hat{R}$ and $n_{\mathrm{eff}}$ values averaged over all elements of the matrix parameter $Y$. }
\label{tab:rhat_neff}
\end{table*}

\begin{figure}[h]
\centering
\vspace{.1in}
\includegraphics[width=0.8\textwidth]{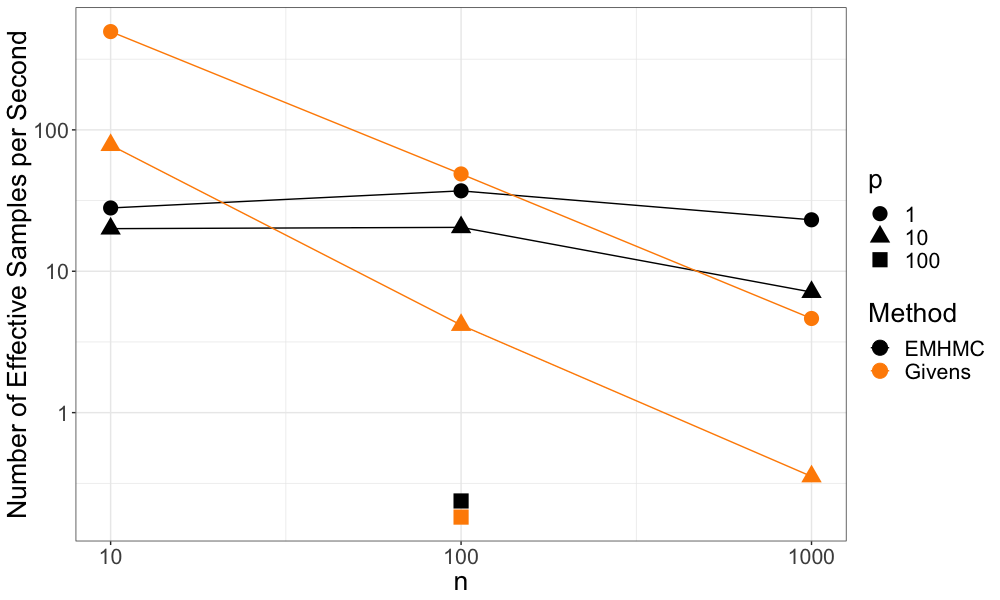}
\vspace{.05in}
\caption{For small values of $n$ the Givens representation approach in Stan produces more effective samplers per second, while for larger values GMC scales better since the primary cost of the matrix exponential remains constant.}
\label{fig:scaling}
\end{figure}

\noindent As mentioned in Section \ref{measureGivens}, to uniformly sample the Stiefel manifold in the Givens representation, the change-of-measure term, Equation \ref{eq:final_change_of_measure}, must be computed as part of the density. Meanwhile, uniform sampling over the Stiefel manifold is achieved in GMC simply using a constant density because the method uses the canonical matrix coordinates. However, as mentioned in section \ref{scaling}, this comes at the cost of an expensive HMC update to ensure that the updated parameter still satisfies the constraints. In practice, we find that GMC scales better as $n$ is increased, although the approach using the Givens representation in Stan remains competitive (Figure \ref{fig:scaling}).

\subsection{Probabilistic Principle Component Analysis (PPCA)}
\begin{figure}[h]
\centering
\vspace{.1in}
\includegraphics[width=0.5\textwidth]{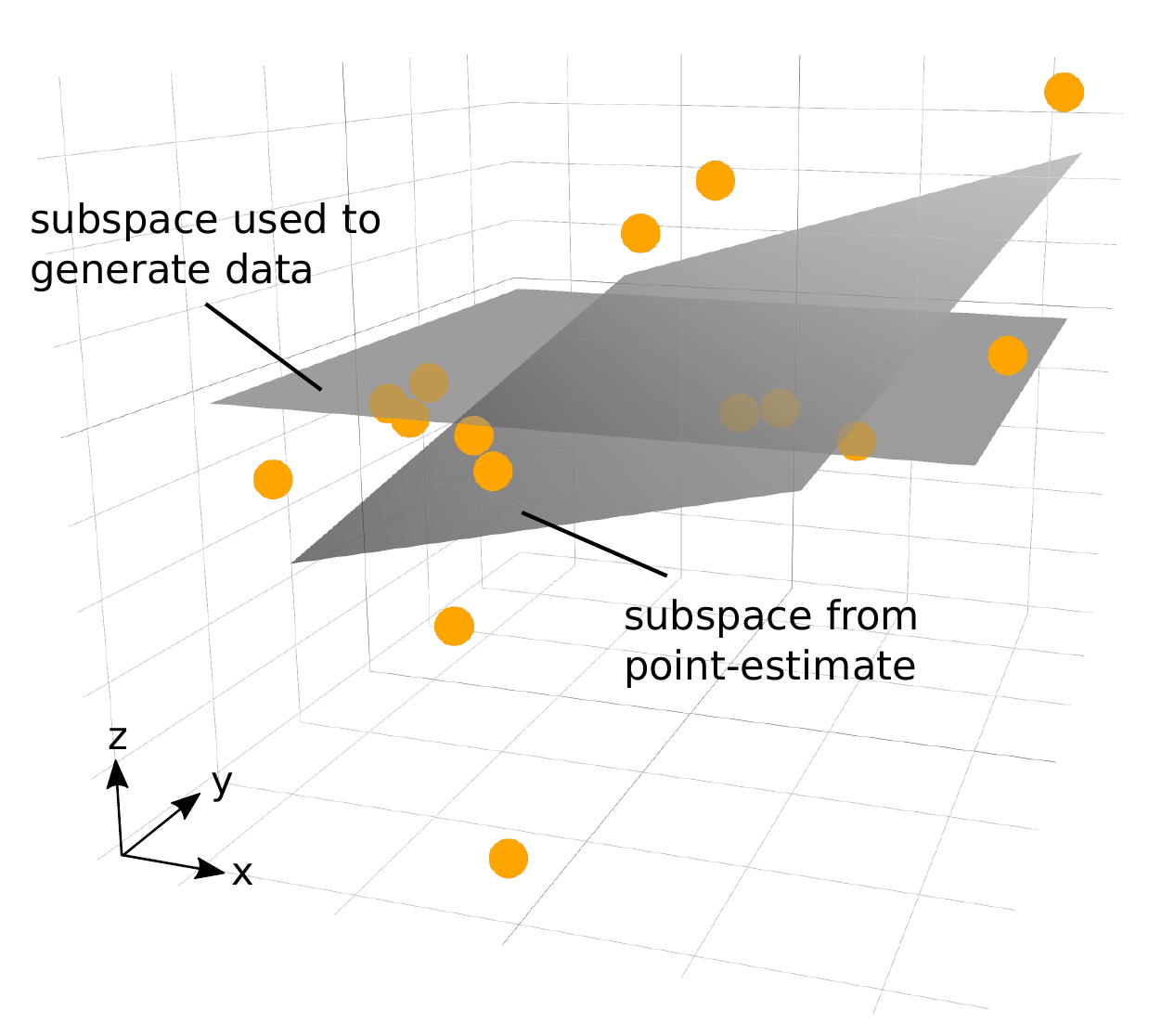}
\vspace{.05in}
\caption{Three dimensional data generated from Equation \ref{eq:PpcaGenerativeProcess} with $p = 2, W = I_{3,2}, \Lambda_{11} = 2, \Lambda_{22} = 1$ and $\sigma = 1$. The maximum-a-posteriori (MAP) estimate of PPCA corresponds to a single orthogonal matrix in the Stiefel Manifold that is closest, in terms of average squared distance, to the set of points. When there are few data points relative to the size of the matrix, this point estimate can often have high variance.}
\label{fig:MleSubspaceEstimate}
\end{figure}

Factor Analysis (FA) and PPCA \citep{tipping1999probabilistic} posit a probabilistic generative model where high-dimensional data is determined by a linear function of some low-dimensional latent state \cite[Chapt.~12]{murphy2012machine}. Geometrically, for a three-dimensional set of points forming a flat pancake-like cloud, the orthogonal matrix corresponding to this linear function can be thought of as a $2$-frame that aligns with this cloud (Figure \ref{fig:MleSubspaceEstimate}). Formally, PPCA posits the following generative process for how a sequence of high-dimensional data vectors $\mathbf{x}_i \in \mathbb{R}^n$, $i = 1, \cdots, N$ arise from some low dimensional latent representations $\mathbf{z}_i \in \mathbb{R}^p$ ($p < n$):

\begin{eqnarray}
\label{eq:PpcaGenerativeProcess}
\mb{z}_i &\sim& \mathcal{N}_p(0, I) \nonumber\\
\mb{x}_i | \mb{z}_i, W, \Lambda, \sigma^2 &\sim& \mathcal{N}_n(W \Lambda \mb{z}_i, \sigma^2 I).
\end{eqnarray}

\noindent To ensure identifiability, $W$ is constrained to be an orthogonal $n \times p$ matrix while $\Lambda$ is a diagonal matrix with positive, ordered elements. Because $\mb{x}_i$ is a linear transformation of a multivariate Gaussian, $\mb{z}_i$ can be integrated out of the model \ref{eq:PpcaGenerativeProcess} to yield the simplified formulation

\begin{eqnarray}
\label{eq:PpcaSimplifiedModel}
\mb{x}_i | W, \Lambda, \sigma^2 &\sim& \mathcal{N}_n(0, \textbf{C}).
\end{eqnarray}

\noindent where $\textbf{C} := W \Lambda^2 W^T + \sigma^2$ I \citep{murphy2012machine}. Letting $\hat{\Sigma} := (1/N) \sum_{i=1}^N \mb{x}_i \mb{x}_i^T$ denote the empirical covariance matrix, this yields the simplified PPCA likelihood

\begin{eqnarray}
p(\mb{x}_1, \cdots, \mb{x}_N | W, \Lambda, \sigma^2) &=& -\frac{N}{2} \ln |\textbf{C}| - \frac{1}{2} \sum_{i=1}^N \mb{x}_i^T \textbf{C}^{-1} \mb{x}_i\\
&=& -\frac{N}{2} \ln |\textbf{C}| - \frac{N}{2} \mathrm{tr} (\textbf{C}^{-1} \hat{\Sigma}).
\label{eq:ppca_likelihood}
\end{eqnarray}

\noindent Traditional PCA corresponds to the closed-form maximum likelihood estimator for $W$ in the limit as $\sigma^2 \to 0$,  providing no measure of uncertainty for this point-estimate. Furthermore, for more elaborate models, the analytical form of the maximum-likelihood estimator is rarely known. Sampling the posterior of a model both provides a measure of uncertainty for parameter estimates and is possible even for more elaborate models.

\noindent We used the Givens representation in Stan to sample the posterior distribution of the parameters in model \ref{eq:PpcaSimplifiedModel} from a simulated dataset with $n = 50$ and $p = 3$. For  $\Lambda$ and $\sigma^2$ we chose uniform priors over the positive real line and for $W$ we chose a uniform prior over the Stiefel manifold yielding the unnormalized posterior density

\begin{eqnarray}
p(W, \Lambda, \sigma^2 | \mb{x}_1, \cdots, \mb{x}_N) \propto p(\mb{x}_1, \cdots, \mb{x}_N | W, \Lambda, \sigma^2),
\end{eqnarray}

\noindent or in the Givens representation

\begin{eqnarray}
\label{eq:ppca_density_givens}
p(\Theta, \Lambda, \sigma^2 | \mb{x}_1, \cdots, \mb{x}_N) \propto p(\mb{x}_1, \cdots, \mb{x}_N | W(\Theta), \Lambda, \sigma^2)\, |J_{Y(\Theta)}(\Theta)|.
\end{eqnarray}

\noindent The latter term comes from Equation \ref{eq:final_change_of_measure}.

\begin{figure}[h]
\centering
\vspace{.1in}
\includegraphics[width=0.7\textwidth]{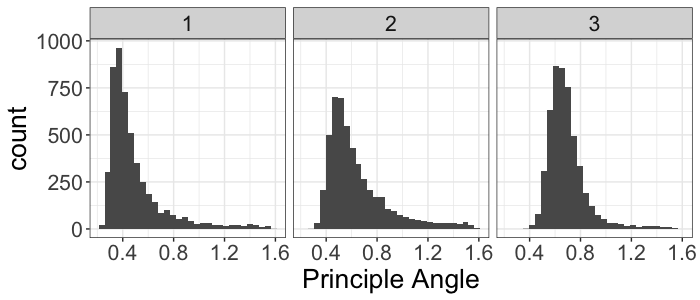}
\vspace{.05in}
\caption{Histograms of the principal angles (in radians) between posterior samples of $W$ and the first three eigenvectors of $\hat{\Sigma}$.}
\label{fig:ppca50_principal_angle}
\end{figure}

\noindent For the true value of the parameters we used the settings used by \citet{jauch2018random} in their experiments section: $\Lambda^2 = \mathrm{diag}(5, 3, 1.5)$, $\sigma^2 = 1$, and $W$ drawn uniformly from $V_{3, 50}$. We took 10,000 samples using Stan's NUTS algorithm with default settings. Table \ref{tab:ppca50} shows the posterior quantiles along with $\hat{R}$ and $n_{\mathrm{eff}}$ values for $\Lambda^2$ and $\sigma^2$. Like \citet{jauch2018random}, we plot histograms of the principal angle,

\begin{equation}
\phi_j = \mathrm{arccos}(E_j^T W_j),\, j=1,2,3
\end{equation}

\noindent between the columns, $W_j$, of posterior draws of $W$ and the columns of the first three eigenvectors of $\hat{\Sigma}$, $E_j$ (Figure \ref{fig:ppca50_principal_angle}).

\begin{table*}
\begin{tabular}{|c||ccccccc|}
\hline
Parameter & 2.5\% & 25\% & 50\% & 75\% & 97.5\% &  $\hat{R}$ & $n_{\mathrm{eff}}$\\
\hline
\hline
$\Lambda_1^2$  & 3.98 & 4.87 & 5.46 & 6.14 & 7.74 & 1.0 & 3,313\\
$\Lambda_2^2$  & 2.48 & 3.16 & 3.61 & 4.09 & 5.10 & 1.0 & 848\\
$\Lambda_2^2$  & 1.21 & 1.76 & 2.07 & 2.44 & 3.20 & 1.0 & 1,340 \\
$\sigma^2$ & 0.99 & 1.01 & 1.03 & 1.04 & 1.07 & 1.0 & 5,374\\
\hline
\end{tabular}
\caption{Posterior quantiles, $\hat{R}$, and $n_{\mathrm{eff}}$ values for $\Lambda^2$ and $\sigma^2$ computed over 10,000 posterior draws.}
\label{tab:ppca50}
\end{table*}

\begin{figure}[h]
\centering
\vspace{.1in}
\includegraphics[width=0.9\textwidth]{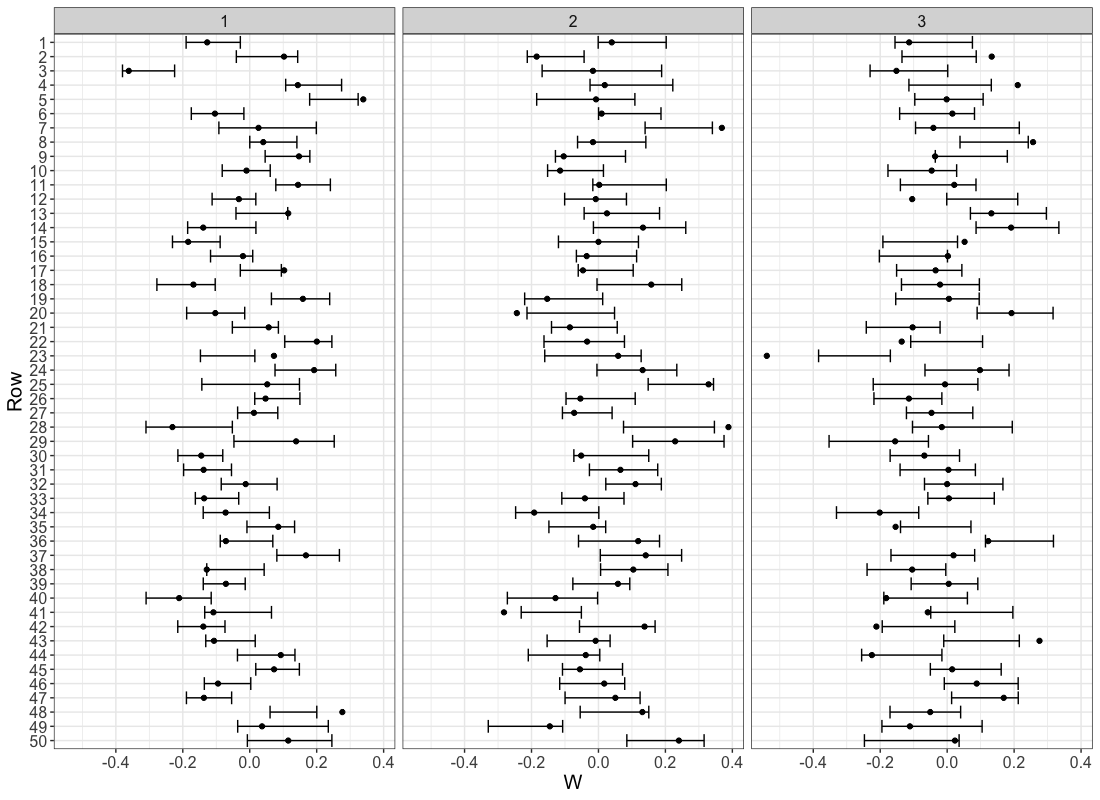}
\vspace{.05in}
\caption{True values of $W$ used in the simulation along with 90\% credible intervals computed using draws of the posterior. Each facet corresponds to one of the three columns of $W$}
\label{fig:ppca50_coverage}
\end{figure}

\noindent We also plot the true values of $W$ used in the simulation along with the 90\% credible intervals, computed from posterior samples, of the marginal posterior distributions of the elements of $W$ (Figure \ref{fig:ppca50_coverage}). 

\subsection{Sparse PPCA}
To illustrate the utility of placing priors over the angle parameters, $\Theta$, of the Givens representation, we fit a PPCA model with sparse priors over $\Theta$ to simulated data generated from \ref{eq:PpcaGenerativeProcess} with the same parameter settings as in the previous example, but with $W$ replaced with a sparse matrix. To generate a sparse orthogonal matrix, we drew an orthogonal matrix uniformly from $V_{3,50}$, converted the result to the Givens representation, randomly set each angle to zero with probability 0.8, then converted the result back to the canonical representation. The result was an orthogonal $V_{3,50}$ matrix, $W$, with 85\% of its elements equal to zero.

\noindent For our model, we used the standard PPCA likelihood \ref{eq:ppca_likelihood} with uniform priors over $\Lambda^2$ and $\sigma^2$, but rather than placing a prior over $\Theta$ to make $W$ uniform over the Stiefel manifold a-priori, we set $\Theta$ to follow the regularized horseshoe prior of \cite{piironen2017sparsity}, a commonly used sparsity-inducing prior. Formally, we set

\begin{eqnarray}
\theta_{ij} &\sim& \mathrm{TruncatedNormal}(0, \tau^2 \tilde{\lambda}_{ij}^2),\; \tilde{\lambda}_{ij}^2 = \frac{c^2 \lambda_{ij}^2}{c^2 + \tau^2 \lambda_{ij}^2}\\
\lambda_{ij} &\sim& \mathrm{Half Cauchy}(0,1)\nonumber \\
\tau &\sim& \mathrm{Half Cauchy}(0,\tau_0)\nonumber \\
c^2 &\sim& \mathrm{Inverse Gamma}(\nu/2, \nu s^2/2)\nonumber
\end{eqnarray}

\noindent with hyper-parameters set to $\tau_0 = 0.01$, $\nu = 10$, and $s = \pi/4$ following the guidelines of \cite{piironen2017sparsity}. We took 10,000 posterior draws in Stan using the resulting unnormalized posterior density

\begin{eqnarray}
\label{eq:ppca_sparse_density_givens}
p(\Theta, \Lambda, \sigma^2 | \mb{x}_1, \cdots, \mb{x}_N) \propto p(\mb{x}_1, \cdots, \mb{x}_N | W(\Theta), \Lambda, \sigma^2)\, p(\Theta, \lambda, \tau, c^2).
\end{eqnarray}

\begin{figure}[h]
\centering
\vspace{.1in}
\includegraphics[width=0.99\textwidth]{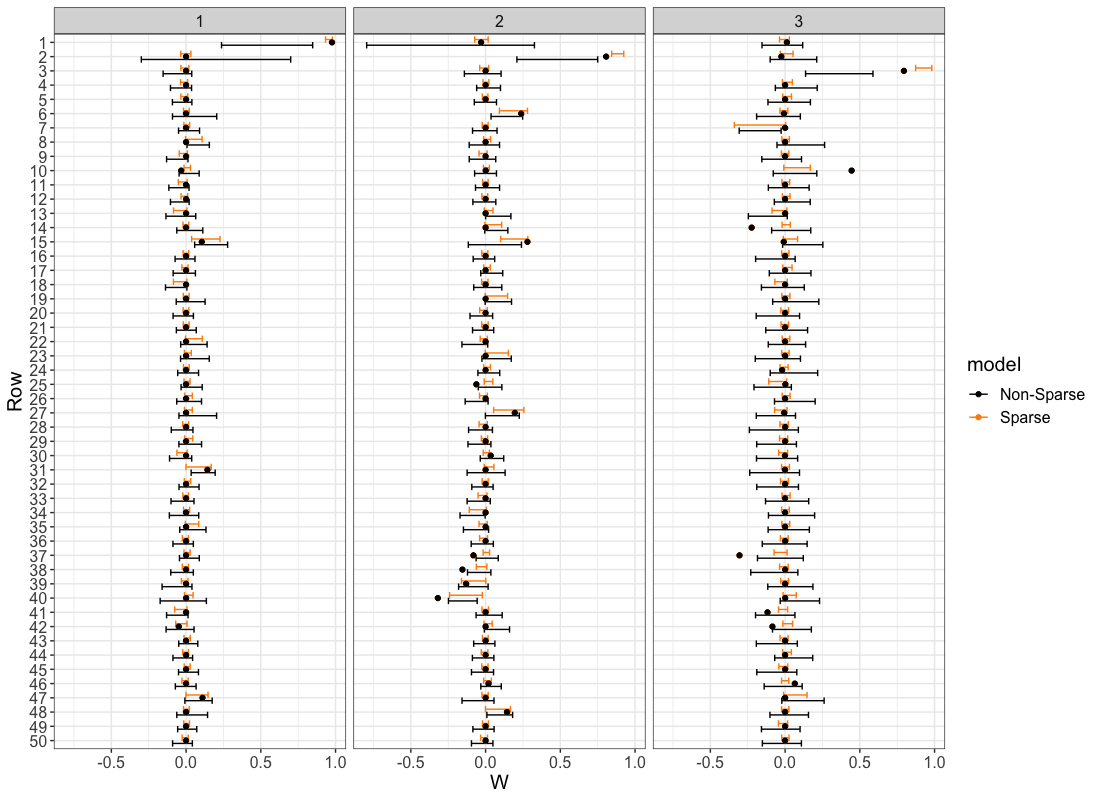}
\vspace{.05in}
\caption{True values of $W$ used in the simulation along with 80\% posterior credible intervals computed using 10,000 draws of the posterior from the sparse and non-sparse models, respectively. Compared to the non-sparse model, the posterior distribution of the sparse model places much more posterior mass close to zero for values that are truly zero, while concentrating mass away from zero for truly non-zero values.}
\label{fig:ppca50_coverage_sparse}
\end{figure}

\noindent Table \ref{tab:ppca50_sparse} shows a posterior summary for the sparse model versus the non-sparse model corresponding to the density \ref{eq:ppca_density_givens}. While the marginal posterior distributions of $\Lambda^2$ and $\sigma^2$ are similar for both models, the sparse model expectedly results in a much sparser posterior distribution over $\Theta$ and thus $W$ (Figure \ref{fig:ppca50_coverage_sparse}). In particular, for elements of $W$ that are truly zero, the marginal posterior distributions of the sparse model tend to concentrate much closer to zero, while for truly non-zero elements, the sparse model is able to concentrate posterior mass away from zero.

\begin{table*}
\begin{tabular}{|cc||ccccccc|}
\hline
Model & Parameter & 2.5\% & 25\% & 50\% & 75\% & 97.5\% &  $\hat{R}$ & $n_{\mathrm{eff}}$\\
\hline
\hline
Non-Sparse & $\Lambda_1^2$  & 3.69 & 4.48 & 4.95 & 5.52 & 6.88 & 1.0 & 3,938\\
Non-Sparse & $\Lambda_2^2$  & 2.66 & 3.38 & 3.80 & 4.25 & 5.16 & 1.0 & 1,828\\
Non-Sparse & $\Lambda_2^2$  & 0.18 & 0.94 & 1.27 & 1.62 & 2.39 & 1.0 & 366 \\
Non-Sparse &  $\sigma^2$ & 0.97 & 1.00 & 1.01 & 1.03 & 1.06 & 1.0 & 1,421\\
\hline
Sparse & $\Lambda_1^2$  & 3.71 & 4.48 & 4.97 & 5.58 & 6.91 & 1.0 & 5,425\\
Sparse & $\Lambda_2^2$  & 2.67 & 3.31 & 3.70 & 4.13 & 5.01 & 1.0 & 4,952\\
Sparse & $\Lambda_2^2$  & 0.78 & 1.15 & 1.38 & 1.63 & 2.24 & 1.0 & 4,779 \\
Sparse &  $\sigma^2$ & 0.97 & 1.00 & 1.01 & 1.03 & 1.05 & 1.0 & 7,191\\
\hline
\end{tabular}
\caption{$\hat{R}$ and $n_{\mathrm{eff}}$ values averaged over all elements of the matrix parameter $Y$. }
\label{tab:ppca50_sparse}
\end{table*}

\subsection{The Network Eigenmodel}
We used the Givens representation to sample from the posterior distribution of the network eigenmodel of \cite{hoff2009simulation} which was also illustrated in \cite{byrne2013geodesic}. We compared the results obtained from using the Givens representation in Stan to results obtained from GMC. The data used in those works and originally described in \cite{butland2005interaction} consists of a symmetric graph matrix, $Y$, of dimension 270 $\times$ 270. However, for our experiments we use a smaller 230 $\times$ 230 version of the dataset as we were unable to find access to the larger version. The version we used is freely available in the R package \textit{eigenmodel}. For our GMC experiments we used the same tuning parameters for GMC as \cite{byrne2013geodesic}.

\noindent The graph matrix encodes whether the proteins in a protein network of size $n=230$ interact with one another. The probability of a connection between all combinations of proteins can be described by the lower-triangular portion of a symmetric matrix of probabilities, however, the network eigenmodel uses a much lower dimensional representation to represent this connectivity matrix. Specifically, given an orthogonal matrix $U$, a diagonal matrix $\Lambda$, and a scalar $c$, then letting $\Phi(\cdot)$ represent the probit link function, the model is described as follows:

\begin{eqnarray}
c &\sim& \mathcal{N}(0, 10^2)\\
\Lambda_i &\sim& \mathcal{N}(0, n),\, \forall i\\
Y_{ij} &\sim& \mathrm{Bernoulli} \left(\Phi ([U \Lambda U^T]_{ij} + c) \right),\, \forall i > j.
\end{eqnarray}

\noindent For $U$ we specified a uniform prior over the Stiefel manifold, which again in the Givens representation corresponds to a prior density that is the absolute value of the change-of-measure term.

\begin{figure}[h]
\centering
\vspace{.1in}
\includegraphics[width=0.8\textwidth]{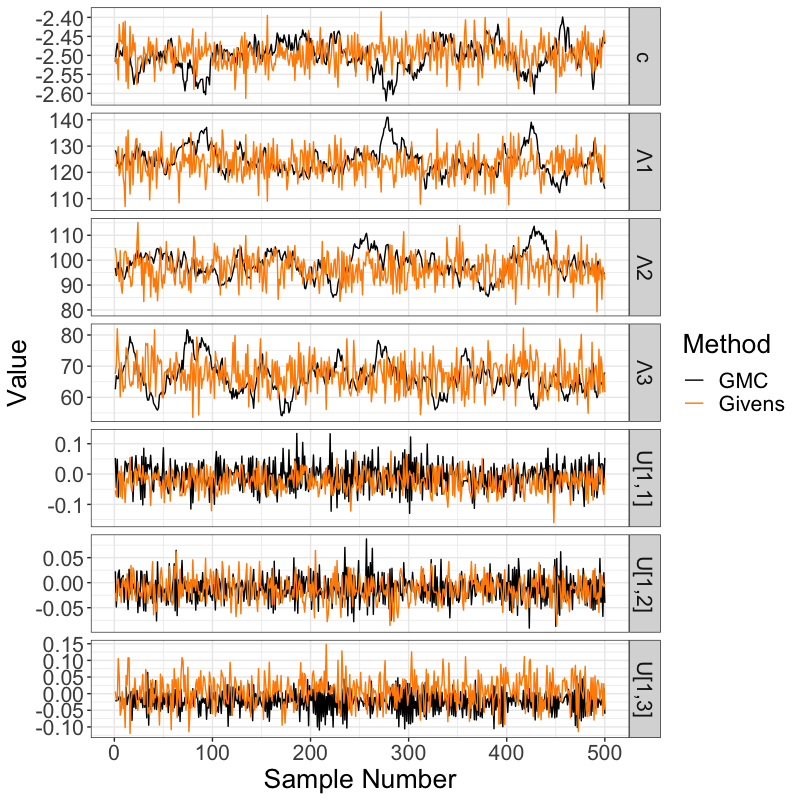}
\vspace{.05in}
\caption{Traceplots of samples from the Givens representation implementation in Stan and GMC. For brevity, only the top three elements of $U$ are shown.}
\label{fig:eigennetwork_traceplots}
\end{figure}

\noindent The Stan implementation using the Givens representation took approximately 300 seconds to collect 1000 samples, 500 of which were warmup. In contrast, GMC took 812 seconds to run the same 1000 samples using the hyperparameter values specified in  \cite{byrne2013geodesic}. Figure \ref{fig:eigennetwork_traceplots} compares traceplots for $c, \Lambda,$ and the elements of the top row $U$ for the 500 post warmup samples from each sampler. Computed $\hat{R}$ and $n_{\mathrm{eff}}$ for these parameters are shown in Table \ref{tab:rhat_neff_eigennetwork}.

\begin{table*}
\begin{tabular}{|c||cc|cc|}
\hline
 & GMC & & Givens &\\
\hline
Parameter & $\hat{R}$ & $n_{\mathrm{eff}}$ & $\hat{R}$ & $n_{\mathrm{eff}}$\\
\hline
\hline
$c$ & 1.00 & 22 & 1.00 & 496\\
$\Lambda_1$ & 1.00 & 19 & 1.00 & 500\\
$\Lambda_2$ & 1.00 & 23 & 1.00 & 500\\
$\Lambda_3$ & 1.10 & 18 & 1.00 & 500\\
$U[1,1]$ & 1.01 & 500 & 1.00 & 500\\
$U[2,1]$ & 1.00 & 500 & 1.00 & 500\\
$U[3,1]$ & 1.02 & 500 & 1.00 & 500\\
\hline
\end{tabular}
\caption{$\hat{R}$ and $n_{\mathrm{eff}}$ values for the parameters in the network eigenmodel. For brevity, only three of the matrix parameters are shown.}
\label{tab:rhat_neff_eigennetwork}
\end{table*}

\section{Discussion}\label{discussion}
We have shown how the Givens representation can be used to develop sampling methods for distributions over the space of orthogonal matrices. We developed approaches for grappling with issues posed by the Stiefel manifold topology and metric. We showed how an auxiliary parameter approach and analytic results for the density measure adjustment terms can be used to develop efficient computational methods. We also showed how our Givens representation approach can be used to specify distributions over the space of orthogonal matrices. We then demonstrated in practice our methods on several examples making comparisons with other approaches.  We expect our introduced Givens representation methods to be applicable to a wide-class of statistical models with orthogonal matrix parameters and to help further facilitate their use in modern probabilistic programming frameworks.

\section{Acknowledgements}\label{acknowledgements}
\noindent Research reported in this publication was performed by the Systems Biology Coagulopathy of Trauma Program of the US Army Medical Research and Materiel Command under award number W911QY-15-C-0026. The author P.J.A acknowledges support from research grant DOE ASCR CM4 DE-SC0009254, DOE ASCR PhILMS DE-SC0019246, and NSF DMS - 1616353.

\appendix
\section{Deriving the Change-of-Measure Term}
We derive the simplified form (Expression \ref{eq:final_change_of_measure}) of the differential form (Expression \ref{eq:WedgeForm}). We point out that \cite{khatri1977mises} provide a similar expression for a slightly different representation, but do not offer a derivation.

\noindent We start with the determinant of the matrix form of the change-of-measure term from Expression \ref{eq:measure_matrix_form} (reproduced below):

\begin{equation}
\begin{pmatrix}
G_{2:n}^T J_{Y_1(\Theta)}(\Theta)\\
G_{3:n}^T J_{Y_2(\Theta)}(\Theta)\\
\vdots\\
G_{p:n}^T J_{Y_p(\Theta)}(\Theta)
\end{pmatrix}
\end{equation}

\noindent For $l = 1, \cdots, n$, we define the following shorthand notation

\begin{equation}
\partial_{i,i+l} Y_k := \frac{\partial}{\partial \theta_{i,i+l}} Y_k
\end{equation}

\noindent and

\begin{equation}
\partial_{i} Y_k
:=
\begin{pmatrix}
\partial_{i,i+1} Y_k & \partial_{i,i+2} Y_k & \cdots & \partial_{in} Y_k.
\end{pmatrix}
\end{equation}

\noindent In the new notation Equation can be written in the following block matrix form:

\begin{equation}
\label{eq:matrix_blockform}
\begin{pmatrix}
G_{2:n}^T \partial_{1} Y_1 &G_{2:n}^T \partial_{2} Y_1 & \cdots & G_{2:n}^T \partial_{p} Y_1\\
G_{3:n}^T \partial_{1} Y_2 &G_{3:n}^T \partial_{2} Y_2 & \cdots & G_{3:n}^T \partial_{p} Y_2\\
\vdots & \vdots & \ddots & \vdots\\
G_{p:n}^T \partial_{1} Y_p &G_{p:n}^T \partial_{2} Y_p & \cdots & G_{p:n}^T \partial_{p} Y_p\\
\end{pmatrix}.
\end{equation}

\noindent Note that the block matrices above the diagonal are all zero.  This can be seen by observing that the rotations in the Givens representation involving elements greater than $i$ will not affect $e_i$, i.e. letting $R_i := R_{i,i+1} \cdots R_{in}$,

\begin{eqnarray}
Y_i = R_1 R_2 \cdots R_p e_i = R_1 \cdots R_i e_i.
\end{eqnarray}

\noindent Thus for $j > i$, $\partial_j Y_i = 0$ and the determinant of Expression \ref{eq:matrix_blockform} simplifies to the product of the determinant of the matrices on the diagonal i.e. the following expression:

\begin{equation}
\label{eq:det_of_blocks}
\prod_{i=1}^p \det \left( G_{i+1:n}^T \partial_{i} Y_i \right).
\end{equation}

\subsection{Simplifying Diagonal Block Terms}
Let $I_{i}$ denote the first $i$ columns of the $n \times n$ identity matrix and let $I_{-i}$ represent the last $n-i$ columns. The term $G_{i+1:n}^T$ in expression \ref{eq:det_of_blocks} can be written as

\begin{equation}
G_{i+1:n}^T = I_{-i}^T G^T = I_{-i}^T R_p^T \cdots R_1^T.
\end{equation}

\noindent To simplify the diagonal block determinant terms in Expression \ref{eq:det_of_blocks} we take advantage of the following fact

\begin{eqnarray}
\det \left( G_{i+1:n}^T \partial_i Y_i  \right)  &=& \det \left( I_{-i}^T R_p^T \cdots R_1^T \right) =  \det\left( I_{-i}^T R_i^T \cdots R_1^T \partial_i Y_i \right).
\end{eqnarray}

\noindent In other words, the terms $R_p^T \cdots R_{i+1}^T$ have no effect on the determinant. This can be seen by first separating terms so that

\begin{eqnarray}
\det\left(G_{i+1:n}^T \partial_{i} Y_i \right) &=& \det\left( \underbrace{I_{-i}^T}_{(n-i) \times n} R_p^T \cdots R_1^T \underbrace{\partial_i Y_i}_{n \times (n-i)} \right)\\
&=& \det\left(
I_{-i}^T
\left[ R_p^T \cdots R_{i+1}^T\right] \left[ R_i^T \cdots R_1^T \partial_i Y_i \right] \right),
\end{eqnarray}

\noindent and then noticing that $R_{i+1} \cdots R_p$ affects only the first $i$ columns of the identity matrix so 

\begin{eqnarray}
I_{-i}^T
\left[ R_p^T \cdots R_{i+1}^T\right]  &=& \left( R_{i+1} \cdots R_p\, I_{-i} \right)^T = \left( I_{-i} \right)^T.
\end{eqnarray}

\noindent Thus Expression \ref{eq:det_of_blocks} is equivalent to

\begin{equation}
\label{eq:simplified_determinant}
\prod_{i=1}^p \det \left( I_{-i}^T R_i^T \cdots R_1^T \partial_{i} Y_i \right).
\end{equation}

\noindent Now consider the $k,l$ element of the $(n-i) \times (n-i)$ block matrix $I_{-i}^T R_i^T \cdots R_1^T \partial_{i} Y_i $. This can be written as 

\begin{eqnarray}
\label{eq:kl_element}
e_{i+k}^T R_i^T \cdots R_1^T \partial_{i,i+l} Y_i &=&  e_{i+k}^T R_i^T \cdots R_1^T \partial_{i,i+l} (R_1 \cdots R_i e_i)\nonumber \\ \nonumber
&=&  e_{i+k}^T R_i^T \cdots R_1^T R_1 \cdots R_{i-1} (\partial_{i,i+l} R_i e_i)\nonumber \\ 
&=&  e_{i+k}^T R_i^T  (\partial_{i,i+l} R_i e_i).
\end{eqnarray}

\noindent Since  $e_{i+k}^T R_i^T R_i e_i =0$, taking the derivatives of both sides and applying the product rule yields

\begin{eqnarray}
\label{eq:kl_element_2}
&&\partial_{i,i+l} (e_{i+k}^T R_i^T R_i e_i) = \partial_{i,i+l} 0\nonumber \\
&\Rightarrow& (\partial_{i,i+l} e_{i+k}^T R_i^T) R_i e_i + e_{i+k}^T R_i^T ( \partial_{i,i+l}R_i e_i) = 0\nonumber \\
&\Rightarrow& e_{i+k}^T R_i^T  (\partial_{i,i+l}R_i e_i) = -(\partial_{i,i+l} e_{i+k}^T R_i^T) R_i e_i.
\end{eqnarray}

\noindent Combining expression \ref{eq:kl_element_2} this fact with expression \ref{eq:kl_element}, the expression for the $k,l$ element of $I_{-i}^T R_i^T \cdots R_1^T \partial_{i} Y_i $ becomes $-(\partial_{i,i+l} e_{i+k}^T R_i^T) R_i e_i$.

\noindent However, note that

\begin{eqnarray}
e_{i+k}^T R_i^T &=&  e_{i+k}^T R_{in}^T \cdots R_{i,i+1}^T = e_{i+k}^T R_{i,i+k}^T \cdots R_{i,i+1}^T,
\end{eqnarray}

\noindent and the partial derivative of this expression with respect to $i,i+l$ is zero when $k > l$. Thus it is apparent that $I_{-i}^T R_i^T \cdots R_1^T \partial_{i} Y_i $ contains zeros above the diagonal and that $\det \left( I_{-i}^T R_i^T \cdots R_1^T \partial_{i} Y_i \right)$ is simply the product of the diagonal elements of the matrix.

\subsection{Diagonal Elements of the Block Matrices}
To obtain the diagonal terms of the block matrices, we directly compute $-\partial_{i,i+l} e_{i+k}^T R_i^T$ for $l=k$, $R_i e_i$, and their inner-product. Defining $D_{ij} := \partial_{ij} R_{ij}$,

\begin{eqnarray}
-\partial_{i,i+k} R_i e_{i+k} &=&   -\partial_{i,i+k} (R_{i,i+1} \cdots R_{i,i+k} e_{i+k}) \\
&=& -R_{i,i+1} \cdots R_{i,i+k-1} D_{i,i+k} e_{i+k} \\
\\
&=&
R_{i,i+1} \cdots R_{i,i+k-1}
\begin{pmatrix}
0\\
\vdots\\
0\\
\cos \theta_{i,i+k}\\
0\\
\vdots\\
0\\
\sin \theta_{i,i+k}\\
0\\
\vdots\\
0
\end{pmatrix}
\\
&=&
R_{i,i+1} \cdots R_{i,i+k-2}
\begin{pmatrix}
0\\
\vdots\\
0\\
\cos \theta_{i,i+k-1} \cos \theta_{i,i+k}\\
0\\
\vdots\\
0\\
\sin \theta_{i,i+k-1} \cos \theta_{i,i+k}\\
\sin \theta_{i,i+k}\\
0\\
\vdots\\
0
\end{pmatrix}
\\
&=&
\begin{pmatrix}
0\\
\vdots\\
0\\
\cos \theta_{i,i+1} \cos \theta_{i,i+2} \cdots \cos \theta_{i,i+k-1} \cos \theta_{i,i+k}\\
\sin \theta_{i,i+1} \cos \theta_{i,i+2} \cdots \cos \theta_{i,i+k-1} \cos \theta_{i,i+k}\\
\vdots\\
\sin \theta_{i,i+k-1} \cos \theta_{i,i+k}\\
\sin \theta_{i,i+k}\\
0\\
\vdots\\
0
\end{pmatrix}
\\
\end{eqnarray}

\noindent which is zero up to the $i$th spot. After the $i+k$th spot,

\begin{eqnarray}
R_i e_i &=&   R_{i,i+1} \cdots R_{in} e_i \\
\\
&=&
\begin{pmatrix}
0\\
\vdots\\
0\\
\cos \theta_{i,i+1} \cos \theta_{i,i+2} \cdots \cos \theta_{i,n-1} \cos \theta_{in}\\
\sin \theta_{i,i+1} \cos \theta_{i,i+2} \cdots \cos \theta_{i,n-1} \cos \theta_{in}\\
\vdots\\
\sin \theta_{i,n-1} \cos \theta_{in}\\
\sin \theta_{in}
\end{pmatrix}.
\end{eqnarray}

\noindent Finally, directly computing the inner-product of $-\partial_{i,i+l} e_{i+k}^T R_i^T$ and $R_i e_i$ yields

\begin{eqnarray}
-(\partial_{i,i+l} e_{i+k}^T R_i^T) (R_i e_i)
&=&
\cos^2 \theta_{i,i+1} \cos^2 \theta_{i,i+2}  \cdots \cos^2 \theta_{i,i+k}  \cos \theta_{i,i+k+1} \cdots \cos \theta_{in} \nonumber\\
&+& \sin^2 \theta_{i,i+1} \cos^2 \theta_{i,i+2} \cdots \cos^2 \theta_{i,i+k}  \cos \theta_{i,i+k+1} \cdots \cos \theta_{in} \nonumber\\
&+&  \sin^2 \theta_{i,i+2} \cos^2 \theta_{i,i+3} \cdots \cos^2 \theta_{i,i+k}  \cos \theta_{i,i+k+1} \cdots \cos \theta_{in} \nonumber\\
&\vdots&  \nonumber\\
&+&  \sin^2 \theta_{i,i+k} \cos \theta_{i,i+k+1} \cdots \cos \theta_{in}
 \nonumber\\
&=&
\cos^2 \theta_{i,i+2} \cos^2 \theta_{i,i+3}  \cdots \cos^2 \theta_{i,i+k}  \cos \theta_{i,i+k+1} \cdots \cos \theta_{in}  \nonumber\\
&+&  \sin^2 \theta_{i,i+2} \cos^2 \theta_{i,i+3} \cdots \cos^2 \theta_{i,i+k}  \cos \theta_{i,i+k+1} \cdots \cos \theta_{in}  \nonumber\\
&\vdots&  \nonumber\\
&+&  \sin^2 \theta_{i,i+k} \cos \theta_{i,i+k+1} \cdots \cos \theta_{in}  \nonumber\\
&=& \cdots \nonumber\\
&=& \cos \theta_{i,i+k+1} \cdots \cos \theta_{in}\nonumber\\
&=& \prod_{k=i+1}^n \cos \theta_{ik}.
\end{eqnarray}

\noindent Thus the determinant of the entire block matrix $I_{-i}^T R_i^T \cdots R_1^T \partial_{i} Y_i $ simplifies to

\begin{equation}
\prod_{k=i+1}^n \left( \prod_{j=k+1}^n \cos \theta_{ik} \right) = \prod_{j=i+1}^n \cos^{j-i-1} \theta_{ij}.
\end{equation}

\noindent Combining this with Expression \ref{eq:simplified_determinant} yields

\begin{eqnarray}
\prod_{i=1}^p \det \left( I_{-i}^T R_i^T \cdots R_1^T \partial_{i} Y_i \right) = \prod_{i=1}^p \prod_{j=i+1}^n \cos^{j-i-1} \theta_{ij}.
\end{eqnarray}

\bibliographystyle{ba}
\bibliography{sample}

\end{document}